\documentclass{article}

\PassOptionsToPackage{numbers}{natbib}
\usepackage[preprint]{neurips_2026}

\usepackage{amsmath,amssymb,amsfonts}
\usepackage{algorithmic}
\usepackage{graphicx}
\usepackage{booktabs}
\usepackage{multirow}
\usepackage{float}

\DeclareUnicodeCharacter{202F}{\,}
\usepackage{booktabs} 
\usepackage{siunitx}  
\usepackage{threeparttablex}   
\usepackage{array}
\usepackage{longtable}
\usepackage{tcolorbox}
\tcbuselibrary{listings,breakable}
\usepackage{tcolorbox}
\tcbuselibrary{breakable}
\usepackage{stfloats} 
\usepackage{soul}
\usepackage{tcolorbox}
\sethlcolor{lime}
\usepackage{ltxtable}
\usepackage{tabularx}
\usepackage{xtab}
\usepackage{makecell}
\usepackage{booktabs}
\usepackage{makecell}
\usepackage{rotating}   
\usepackage{algorithm,algorithmic}
\usepackage[colorlinks=true,linkcolor=black,citecolor=black,urlcolor=black,hypertexnames=true]{hyperref}
\usepackage{textcomp}
\usepackage{placeins}

\begin{document}
\title{QuanTiMedAI: Quantum-Enhanced Time-Series Model guided by Agentic AI for Cardiac Arrest Mortality Prediction}


\author{
  Mutasim Fuad Sarker \\
  Dept.\ of ECE \\
  North South University, Dhaka, Bangladesh \\
  \texttt{mutasim.sarker@northsouth.edu}
  \And
  Adiba Rahman Namira \\
  Dept.\ of ECE \\
  North South University, Dhaka, Bangladesh \\
  \texttt{adiba.namira@northsouth.edu}
  \And
  Wafa Binte Alam \\
  Dept.\ of ECE \\
  North South University, Dhaka, Bangladesh \\
  \texttt{wafa.alam@northsouth.edu}
  \AND
  Md Adnan Arefeen \\
  Dept.\ of ECE \\
  North South University, Dhaka, Bangladesh \\
  \texttt{adnan.arefeen@northsouth.edu}
  \And
  Mahzabeen Emu \\
  Dept.\ of ECE \\
  Memorial University, St.\ John's, NL, Canada \\
  \texttt{memu@mun.ca}
  \And
  Sumaiya Tabassum Nimi \\
  Dept.\ of ECE \\
  North South University, Dhaka, Bangladesh \\
  \texttt{sumaiya.nimi@northsouth.edu}
}

\maketitle

\begin{abstract}
Cardiac arrest remains one of the most lethal conditions encountered in intensive care units. Despite the growing availability of electronic health record data, existing mortality prediction studies in this population largely depend on static summaries derived from early admission. Such approaches ignore the temporal progression of physiological deterioration and recovery that unfolds throughout a patient's ICU stay. To address this limitation, we introduce QuanTiMedAI, a quantum-agentic framework developed for cardiac arrest mortality prediction using agentic AI guided quantum enhancement time series model. The proposed system combines an agentic large language model (LLM) for clinically informed feature discovery with a compact quantum recurrent network for temporality aware mortality prediction. Our findings demonstrate that agentic LLM-guided feature selection consistently outperforms conventional feature selection approaches, and the 
proposed quantum architecture achieves competitive predictive performance through nonlinear feature enhancement while keeping the number of parameters very low. Through extensive experimentation on a MIMIC-IV cohort of cardiac arrest patients, QuanTiMedAI's quantum-enhanced architecture attains an AUROC of 0.852 using only 605 parameters, an improvement of approximately 2.9\% over a current state-of-the-art baseline for this task. A structured ablation study systematically validates the contribution of each architectural design choice. These results show that quantum-enhanced sequential modeling can exceed classical recurrent networks while using substantially fewer parameters.

\end{abstract}

\textbf{Keywords:} 
MIMIC-IV, Large Language Model, , Quantum Long Short Term Memory (QLSTM), Timeseries Analysis, Clinical Decision Support.

\section{Introduction}

 Cardiac arrest is one of the most lethal and time-critical emergencies
encountered in hospital medicine. In-hospital cardiac arrest (IHCA)
affects millions of patients worldwide each year, and despite decades
of improvement in resuscitation
protocols~\cite{callaway2014survival, vo2024management}, survival
rates remain persistently poor. A recent nationwide cohort study
reported an incidence of approximately 6 to 7 IHCA events per
1{,}000 hospitalizations~\cite{chang2025longitudinal}, and a
systematic review and meta-analysis across 35 observational studies
confirmed pooled ICU mortality of 74\% and in-hospital mortality
of 82\% in ICU-admitted cardiac arrest
patients~\cite{lim2026systematic}. Early identification of high-risk patients directly shapes treatment decisions, resource allocation, and goals-of-care discussions in a setting where clinical conditions can evolve rapidly~\cite{wei2025machine}.

\begin{figure}[!t]
    \centering
    \includegraphics[width=0.9\linewidth]{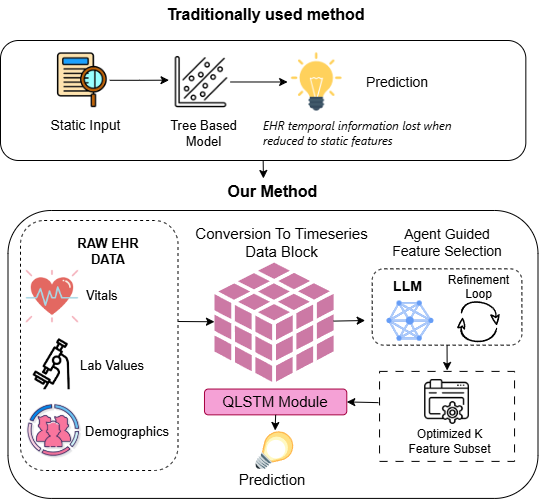}
    \caption{QuanTiMedAI overview: comparison of the traditional static tree-based prediction pipeline against our proposed approach, where raw ICU EHR data is converted to time-series blocks and passed through an agentic Gemma 4 driven feature selection loop before being fed into the QLSTM module for mortality prediction.}
    \label{fig:teaser_Framework}
\end{figure}

The availability of large-scale electronic health record repositories such as MIMIC-IV~\cite{johnson2023mimic} has accelerated data-driven cardiac arrest mortality research, and several groups have applied machine learning to this problem using this
resource~\cite{sun2023prediction, liu2025prediction, li2025prognostic,
jia2026prediction, yu2025benchmarking}. Broader systematic reviews confirm that traditional
machine learning approaches have consistently dominated cardiac arrest
outcome prediction research across the
literature~\cite{wei2025machine, zobeiri2025systematic}. These studies
represent real progress in showing that structured clinical data can
support mortality prediction in this population.

A closer examination of these studies reveals a common methodological trend: to our knowledge most MIMIC-IV–based cardiac arrest mortality prediction models rely primarily on conventional machine learning approaches, including LASSO, XGBoost, and ensemble-based methods.~\cite{sun2023prediction, liu2025prediction, li2025prognostic,
jia2026prediction}. These models each patient as a fixed snapshot with no way to capture
how a patient's condition evolves over time. ICU data is inherently
sequential and vital signs, laboratory values, and treatment events
evolve continuously throughout a patient's stay, with the path of
that change carrying prognostic information that a single static
snapshot simply cannot
capture~\cite{harutyunyan2019multitask, nowroozilarki2021realtime, theis2022process}. Sequential models have been
shown to clearly outperform static approaches on general ICU
populations~\cite{deng2022explainable, zheng2025dynamic, yu2025benchmarking, shickel2018deepehr, nowroozilarki2021realtime, liu2025mstdgnn}, yet to
the best of our knowledge, no prior study has investigated
sequential time-series modeling for cardiac arrest mortality
prediction on MIMIC-IV.

Classical LSTM, while effective for sequential data, requires large
numbers of parameters and is known to struggle with complex nonlinear
temporal dependencies in long
sequences~\cite{hochreiter1997long}. Quantum Long Short-Term Memory
(QLSTM)~\cite{chen2022qlstm} addresses this by replacing classical
gating with variational quantum circuits, offering higher
representational capacity per
parameter~\cite{biamonte2017quantum, abbas2021expressibility}, and
has outperformed classical LSTM across multiple sequential forecasting
domains~\cite{kea2024hybrid, su2025bls, khan2024qlstm}. Systematic
reviews of quantum machine learning in healthcare have consistently
identified sequential clinical EHR data as the most important
underexplored area~\cite{gupta2025systematic, ullah2024quantum},
yet no prior work has applied QLSTM to clinical EHR time-series
for patient outcome prediction~\cite{zhang2026quantum}.

These studies also rely on conventional feature selection methods
such as LASSO regularization and correlation filtering. These approaches are purely data-driven and offer limited capacity to incorporate existing clinical knowledge about cardiac arrest physiology. The
features they select may perform well statistically on a given
training split but are not necessarily grounded in clinical
reasoning~\cite{jeong2024llmfsagent}. For a problem like cardiac
arrest mortality, where the relevant physiology is well understood,
this is a real gap.

Agentic AI offers a promising alternative. Large language models have been shown to encode substantial medical knowledge acquired from scientific literature and clinical corpora~\cite{singhal2023medpalm, yu2025benchmarking}. Recent research further suggests that these models can identify clinically meaningful predictors using domain reasoning rather than relying exclusively on statistical associations~\cite{jeong2024llmfsagent, choi2024featllm}. Unlike conventional prompt-based approaches, an agentic framework enables iterative refinement, allowing the model to evaluate performance, reflect on outcomes, and revise feature selections accordingly~\cite{jeong2024llmfsagent}. For cardiac arrest mortality prediction, where many risk factors are already well established in clinical literature, this feedback-driven process provides a practical mechanism for identifying informative variables beyond what can be achieved through manual selection or statistical filtering alone~\cite{singhal2023medpalm, jeong2024llmfsagent}.

This work proposes \textbf{QuanTiMedAI}, a quantum-agentic
clinical time-series framework that combines agentic LLM-guided
feature selection with a compact quantum recurrent architecture
for in-hospital cardiac arrest mortality prediction on MIMIC-IV,
evaluated against classical LSTM baselines under identical
experimental conditions. To the best of our knowledge, agentic
LLM reasoning for clinical feature discovery and quantum recurrent
modeling for temporal prediction have not previously been brought
together within a single clinical decision-support pipeline. The principal contributions of this
work are as follows:

\begin{itemize}

    \item We propose QuanTiMedAI, a quantum-agentic time-series framework for in-hospital cardiac arrest mortality prediction on MIMIC-IV that integrates agentic LLM-guided feature selection, an engineered clinical severity-score channel, and a compact quantum recurrent network for sequential modeling of ICU data.

    \item We introduce an agentic LLM-guided feature selection pipeline
    that iteratively refines clinical feature subsets through
    performance-guided feedback, incorporating clinical domain knowledge
    via an engineered severity score channel.

    \item We propose a modified five-VQC QLSTM architecture with an input re-injection skip connection that improves AUROC by about 2\% over a parameter-matched classical LSTM (0.852 versus 0.835) on the same cohort, while using only 605 parameters, yielding a lightweight model.

    \item We conduct a rigorous ablation study across six
    architectural variants that isolates the effect of each design choice, showing that the proposed design attains the highest AUROC (0.852) and that the input re-injection mechanism alone accounts for roughly a 1.5\% AUROC gain.

\end{itemize}

\section{Related Work}
\label{sec:related}

\subsection{In-Hospital Cardiac Arrest Mortality Prediction}

The MIMIC-IV database is a publicly available de-identified electronic health
record dataset from Beth Israel Deaconess Medical Center, covering ICU
admissions from 2008 to 2019 and accessible to credentialed researchers via
PhysioNet \cite{johnson2023mimic}. It has been widely used to study
in-hospital mortality prediction in cardiac arrest patients. A previous study
conducted a retrospective study on 1,722 ICU-admitted cardiac arrest patients,
applying LASSO regression and XGBoost to identify independent risk factors
from demographics, comorbidities, vital signs, laboratory results, scoring
systems, and treatment information collected from the first day of ICU
admission. The LASSO nomogram was selected as the final model based on higher
net benefit \cite{sun2023prediction}. Some other studies extended this to a
multicenter setting using both MIMIC-IV and the eICU Collaborative Research
Database, building an ensemble model through soft voting across eight machine
learning algorithms, with the externally validated model deployed as a
clinical web application \cite{liu2025prediction}. A more recent study on
MIMIC-IV applied multiple machine learning algorithms to post-cardiac arrest
patients for 28-day mortality prediction, where XGBoost achieved an AUC-ROC
of 0.89 and dynamic lactate clearance was identified as the primary predictor
\cite{chen2026superior}.

While these studies show that machine learning on structured clinical data
from MIMIC-IV can support cardiac arrest mortality prediction, they all share
the same critical limitation. Every study aggregated clinical variables from
only the first 24 hours of ICU admission into a static feature vector,
discarding the temporal dimension entirely. ICU EHR data is inherently
sequential. Vital signs, laboratory values, scoring systems, medication
events, and output measurements are recorded repeatedly at regular intervals
throughout a patient's stay, and the patterns of change in these measurements
over time carry useful predictive information that a single-timepoint snapshot
cannot capture. Previous studies acknowledged this directly, noting that
traditional machine learning models ignore the time-series characteristics of
ICU data~\cite{nowroozilarki2021realtime}, and applied RNN, GRU, and LSTM with attention mechanisms to 40,083
general ICU patients from MIMIC-IV, achieving an AUC of 0.870 $\pm$ 0.001
for in-hospital mortality prediction \cite{deng2022explainable, yu2025benchmarking}. A past
research similarly demonstrated on MIMIC-III that LSTM-based sequential
models significantly outperform static classifiers when the temporal
structure of clinical measurements is preserved
\cite{harutyunyan2019multitask}.

\begin{table*}[htbp]
\centering
\caption{Comparison of the proposed QuanTiMedAI framework with recent in-hospital mortality prediction studies.}
\label{tab:model_comparison_transposed}
\footnotesize
\setlength{\tabcolsep}{6pt}
\renewcommand{\arraystretch}{1.6}
\begin{tabular}{l c c c c c}
\toprule
\multirow{2}{*}{\textbf{Study}} & \multirow{2}{*}{\makecell{\textbf{CA}\\\textbf{Cohort}}} & \multirow{2}{*}{\makecell{\textbf{Seq.\ 3-D}\\\textbf{Temporal}}} & \multicolumn{2}{c}{\textbf{Model Architecture}} & \multirow{2}{*}{\makecell{\textbf{Agentic}\\\textbf{LLM}}} \\
\cmidrule(lr){4-5}
& & & \textbf{Classical ML} & \textbf{QLSTM} & \\
\midrule
Sun et al.~\cite{sun2023prediction}       & \checkmark & $\times$ & \checkmark~(LASSO)    & $\times$ & $\times$ \\
Liu et al.~\cite{liu2025prediction}       & \checkmark & $\times$ & \checkmark~(Ensemble) & $\times$ & $\times$ \\
Li et al.~\cite{li2025prognostic}         & \checkmark & $\times$ & \checkmark~(XGBoost)  & $\times$ & $\times$ \\
Jia et al.~\cite{jia2026prediction}       & \checkmark & $\times$ & \checkmark~(XGBoost)  & $\times$ & $\times$ \\
Deng et al.~\cite{deng2022explainable}    & $\times$   & \checkmark & \checkmark~(LSTM)   & $\times$ & $\times$ \\
Wang et al.~\cite{wang2026lungcancer}     & $\times$   & $\times$ & \checkmark~(XGBoost)  & $\times$ & $\times$ \\
{\bfseries QuanTiMedAI(Ours)}           & \checkmark & \checkmark & \checkmark~(LSTM)   & \checkmark & \checkmark \\
\bottomrule
\end{tabular}
\end{table*}

Table~\ref{tab:model_comparison_transposed} summarizes these differences
across four recent cardiac arrest mortality studies on MIMIC-IV~\cite{sun2023prediction,
liu2025prediction, li2025prognostic, jia2026prediction} and two broader ICU
mortality studies~\cite{deng2022explainable, wang2026lungcancer} alongside
our proposed method. All prior studies rely solely on classical machine
learning approaches with no sequential temporal representation, no quantum
architecture, and no agentic LLM guided feature selection. Our proposed
method is the only entry that incorporates all of these dimensions
simultaneously, combining sequential three-dimensional temporal modeling,
LSTM and QLSTM architectures, agentic LLM feature selection, and LLM
reasoning for interpretability.

To the best of our knowledge, no prior study identified in our
literature review has applied LSTM or any sequential time-series
deep learning model to in-hospital mortality prediction in cardiac
arrest patients using MIMIC-IV EHR data. All existing cardiac
arrest mortality studies on MIMIC-IV identified in this review have
relied solely on static first-day feature aggregations, leaving how
a patient's condition changes over time entirely unmodeled. Also, even where LSTM has been applied to general ICU
time-series data, it has known limitations such as gradient instability on
long sequences and limited capacity for highly nonlinear temporal dependencies
\cite{hochreiter1997long}, which motivate exploring a more capable
alternative. This is where QLSTM becomes relevant, as discussed in the
following subsection.

\subsection{QLSTM in Time-Series Applications}

QLSTM was introduced as a hybrid quantum-classical architecture that replaces
the classical gating mechanisms of LSTM cells with variational quantum
circuits \cite{chen2022qlstm}. The authors showed that QLSTM successfully
learns temporal sequential data and in certain cases converges faster and
reaches better accuracy than classical LSTM, while its shallow circuit
requirements keep it practical for near-term NISQ devices \cite{chen2022qlstm}.

Since then, QLSTM has been validated across multiple time-series domains and
has consistently outperformed classical LSTM. In financial forecasting, a
study showed that a hybrid QLSTM outperformed classical LSTM on stock price
prediction across multiple evaluation metrics \cite{kea2024hybrid}, and a
study combining broad learning systems with QLSTM for chaotic stock index
forecasting showed consistent improvements over classical LSTM on real
financial datasets \cite{su2025bls}. The pattern across these studies is
clear: on complex nonlinear sequential data, QLSTM performs better than
classical LSTM.

On the healthcare side, multiple systematic reviews have recognized the
potential of quantum machine learning for clinical data. Ullah et al.\
reviewed 49 QML studies in healthcare covering EHR data, ECG, and medical
imaging, and identified sequential clinical EHR data as an important
underexplored direction \cite{ullah2024quantum}. Subsequent studies similarly
identified potential for quantum-enhanced models in clinical prediction tasks
in a systematic review of QML in the biomedical domain
\cite{maheshwari2022quantum}. A study has the most comprehensive review to
date published in \textit{npj Digit.\ Med.}\ screened over 4,915 studies from
2015 to 2024 and found that while QML for digital health is growing rapidly,
rigorous application-specific work on clinical EHR data particularly for
critical care outcome prediction remains scarce \cite{gupta2025systematic}.

There is only one existing application of QLSTM in a biomedical context in
which it was applied to drug discovery on static molecular fingerprint
datasets, achieving ROC-AUC improvements of 3\% to over 6\% over classical
LSTM \cite{zhang2026quantum}. However, this work involves no sequential or
temporal modeling, no patient EHR data, and no clinical outcome prediction.
The temporal modeling capability for which QLSTM was designed was not used
at all.

This leaves two clear and directly related gaps. First, no prior study
identified in our review has applied any sequential time-series model to
in-hospital mortality prediction in cardiac arrest patients on MIMIC-IV. Second, even though QLSTM has consistently outperformed classical LSTM on time-series data in multiple
domains, it has never been applied to sequential clinical EHR data for any
patient outcome prediction task, and the only existing biomedical application
of QLSTM used static data rather than time-series. Given that medical ICU EHR
data is multivariate, hourly, and spans the full patient stay, which is
exactly the kind of sequential data where QLSTM has shown advantages over
LSTM, there is strong motivation to apply it here. To the best of our knowledge, this work is among the first to apply
QLSTM to sequential clinical EHR time-series data for patient outcome
prediction, and among the first to apply sequential deep learning to
in-hospital cardiac arrest mortality prediction on MIMIC-IV.

\begin{table*}[htbp]
\centering
\caption{Baseline patient characteristics stratified by in-hospital mortality.}
\label{tab:baseline_char}

\begin{tabular}{
    @{} 
    l
    c
    c
    c
    c
    @{}
}
\toprule
\textbf{Characteristic} 
& \textbf{Died} 
& \textbf{Survived} 
& \textbf{SMD} 
& \textbf{\textit{p}-value} \\
& \textbf{(n = 1296)} 
& \textbf{(n = 1011)} 
& 
&  \\
\midrule
Age, years                         & 67.8 $\pm$ 16.4 & 64.5 $\pm$ 16.2 &  0.199 & $<0.001$ \\
Male sex                           & 747 (57.6\%)    & 641 (63.4\%)    & -0.118 & 0.006 \\
Hypertension                       & 846 (65.3\%)    & 715 (70.7\%)    & -0.117 & 0.006 \\
Congestive heart failure           & 484 (37.3\%)    & 451 (44.6\%)    & -0.148 & $<0.001$ \\
Myocardial infarction              & 361 (27.9\%)    & 327 (32.3\%)    & -0.098 & 0.022 \\
Diabetes mellitus                  & 473 (36.5\%)    & 361 (35.7\%)    &  0.016 & 0.728 \\
Chronic obstructive pulmonary disease & 295 (22.8\%) & 204 (20.2\%)    &  0.063 & 0.148 \\
Prior stroke                       & 157 (12.1\%)    & 90 (8.9\%)      &  0.105 & 0.016 \\
\bottomrule
\end{tabular}

\vspace{0.5em}
\begin{minipage}{0.95\textwidth}
\footnotesize
\textit{Note.} Continuous variables are reported as mean $\pm$ standard deviation, and categorical variables are reported as count (\%). SMD denotes standardized mean difference. An absolute SMD greater than 0.10 is conventionally considered to indicate meaningful imbalance. \textit{p}-values were obtained using appropriate univariate tests, such as the two-sample \textit{t}-test for continuous variables and the $\chi^2$ test for categorical variables.
\end{minipage}
\end{table*}


\noindent\textbf{Baseline Characteristics Description.}
The cohort comprised 2,307 patients, of whom 1,296 (56.2\%) died during hospitalization. Several baseline characteristics differed significantly between the mortality groups. Patients who died were older (SMD = 0.199, $p<0.001$) and had a higher prevalence of prior stroke (SMD = 0.105, $p=0.016$). In contrast, the survived group had larger proportions of males, as well as higher rates of hypertension, congestive heart failure, and myocardial infarction (all $\lvert\text{SMD}\rvert \geq 0.098$, $p<0.05$). Diabetes and chronic obstructive pulmonary disease were well balanced (SMD $\leq 0.063$, $p > 0.05$). Notably, age, sex, hypertension, CHF, and stroke all exceeded the conventional SMD threshold of 0.10 for potential imbalance, suggesting that these variables should be considered for adjustment in downstream analyses.

\section{Methods}

This section presents our proposed methodological framework called
QuanTiMedAI, designed to evaluate whether agentic LLM-guided
feature selection improves in-hospital mortality prediction in
cardiac arrest ICU patients relative to random feature selection,
and whether replacing classical LSTM with the proposed QLSTM
yields further improvement. Three experimental conditions share
an identical data source, cohort definition, preprocessing
pipeline, training procedure, and evaluation protocol. The
controlled variables are (i) the feature selection strategy
and (ii) the recurrent cell architecture, such that any
observed difference in performance can be attributed to one
of these two factors.

\subsection{Dataset and Cohort}

\subsubsection{Data Source and Inclusion Criteria}
Data were sourced from the MIMIC-IV database~\cite{johnson2023mimic},
a publicly available de-identified electronic health record
repository from Beth Israel Deaconess Medical Center. Patients
were identified by cardiac arrest diagnosis codes, specifically
ICD-9 code 427.5 and ICD-10 codes prefixed with I46. Inclusion
was restricted to adult patients aged $\geq 18$ years at the
time of admission who had a linked ICU stay. For patients with
multiple hospital admissions, only the first admission was
retained, and within each admission, only the first documented
ICU stay was used. The prediction task was defined as binary
in-hospital mortality classification using clinical data
collected exclusively within the first $\tau = 24$ hours of
ICU admission~\cite{yu2025benchmarking}, where $\tau$ denotes the observation window.

\subsubsection{Feature Extraction and Preprocessing}
A total of $F$ clinical features were extracted spanning
multiple physiological domains, including vital signs, severity
scores, haematology, coagulation, renal function, metabolic
and electrolyte panels, arterial blood gas measurements,
cardiac biomarkers, hepatic markers, fluid balance, vasoactive
agent administration, and mechanical ventilation status.
Each feature is designated as either temporal (T) or static
(S). Temporal features are just the ongoing measurements tracked over the observation period, while static features cover the fixed details like a patient's age, sex, and baseline medical history.

Temporal features were binned into $T$ equally spaced
intervals over the observation window $\tau$, where the
bin width is $\tau / T$ hours and
\begin{equation}
    T \in \mathcal{T},
    \label{eq:T}
\end{equation}
with $\mathcal{T}$ denoting the discrete set of candidate
temporal resolutions evaluated in this study. Within each
bin, continuous variables were summarized by their mean
value, urine output was aggregated as a cumulative sum,
and vasoactive agent administration and mechanical ventilation
were encoded as binary indicators denoting active
administration during that interval.

Features with a missingness rate exceeding a predefined
threshold $\rho$ were excluded prior to model training.
Clinically critical features identified as having established
prognostic value in cardiac arrest were retained regardless
of their missingness rate. For the remaining features,
missing values were handled using forward fill along the
temporal axis, with leading missing values back-filled
from the first available observation. All temporal features
were normalized using z-score standardization:
\begin{equation}
    \tilde{x}_{i}(t) =
    \frac{x_{i}(t) - \mu_{i}}{\sigma_{i}},
    \label{eq:zscore}
\end{equation}
where $\mu_{i}$ and $\sigma_{i}$ denote the mean and
standard deviation of feature $i$ computed exclusively
from the training partition and applied to the validation
and test sets to prevent data leakage. The resulting input
for each patient is a three-dimensional tensor of shape
$(T, F_{\text{sel}} + 1)$, where $F_{\text{sel}}$ denotes
the number of selected features and the additional channel
corresponds to the engineered severity score described
in the following subsection.

\begin{table}
\renewcommand{\arraystretch}{1.15}
\scriptsize
\caption{Clinical Features Extracted From MIMIC-IV for 24-Hour ICU Cardiac Arrest Patients}
\label{tab:features}
\centering

\begin{threeparttable}

\begin{tabular}{@{} l >{\raggedright\arraybackslash}p{4.2cm} c @{}}
\toprule
\textbf{Category} &
\textbf{Feature} &
\textbf{Type} \\
\midrule

\multirow{10}{*}{\makecell[l]{\textbf{Vital Signs}}}
  & Heart Rate                                & T \\
  & Systolic Blood Pressure                   & T \\
  & Diastolic Blood Pressure                  & T \\
  & Mean Arterial Pressure                    & T \\
  & Respiratory Rate                          & T \\
  & Body Temperature                          & T \\
  & Peripheral Oxygen Saturation (SpO$_2$)    & T \\
  & Glasgow Coma Scale (Total)\tnote{a}       & T \\
\midrule

\makecell[l]{\textbf{Severity}\\\textbf{Scores}} 
  & SAPS III Score & T \\
\midrule

\multirow{6}{*}{\makecell[l]{\textbf{Haematology}}}
  & Haematocrit                               & T \\
  & Haemoglobin                               & T \\
  & Platelet Count                            & T \\
  & White Blood Cell Count                    & T \\
  & Red Blood Cell Count                      & T \\
  & Red Cell Distribution Width               & T \\
\midrule

\multirow{2}{*}{\makecell[l]{\textbf{Coagu-}\\\textbf{lation}}}
  & Prothrombin Time                          & T \\
  & International Normalised Ratio            & T \\
\midrule

\multirow{2}{*}{\makecell[l]{\textbf{Renal}\\\textbf{Function}}}
  & Serum Creatinine                          & T \\
  & Blood Urea Nitrogen                       & T \\
\midrule

\multirow{10}{*}{\makecell[l]{\textbf{Metabolic /}\\\textbf{Electrolytes}}}
  & Blood Glucose                             & T \\
  & Serum Potassium                           & T \\
  & Serum Sodium                              & T \\
  & Serum Calcium                             & T \\
  & Serum Chloride                            & T \\
  & Anion Gap                                 & T \\
  & Serum Bicarbonate                         & T \\
  & Serum Magnesium                           & T \\
  & Serum Albumin                             & T \\
  & Serum Lactate                             & T \\
\midrule

\multirow{4}{*}{\makecell[l]{\textbf{Arterial}\\\textbf{Blood Gas}}}
  & Arterial pH                               & T \\
  & Partial Pressure of O$_2$                 & T \\
  & Partial Pressure of CO$_2$                & T \\
  & Base Excess                               & T \\
\midrule

\multirow{4}{*}{\makecell[l]{\textbf{Cardiac /}\\\textbf{Hepatic}}}
  & Cardiac Troponin T                        & T \\
  & Creatine Kinase-MB                        & T \\
  & Alanine Aminotransferase                  & T \\
  & Aspartate Aminotransferase                & T \\
\midrule

\textbf{Fluid Balance}
  & Urine Output (hourly sum)                 & T \\
\midrule

\multirow{5}{*}{\makecell[l]{\textbf{Vasoactive}\\\textbf{Agents}\tnote{b}}}
  & Epinephrine                               & T \\
  & Dopamine                                  & T \\
  & Norepinephrine                            & T \\
  & Phenylephrine                             & T \\
  & Dobutamine                                & T \\
\midrule

\textbf{Ventilation}\tnote{b}
  & Mechanical Ventilation                    & T \\
\midrule

\multirow{2}{*}{\textbf{Demographics}}
  & Age at ICU Admission                      & S \\
  & Sex (Male = 1)                            & S \\
\midrule

\multirow{6}{*}{\textbf{Comorbidities}\tnote{b}}
  & Hypertension                              & S \\
  & Congestive Heart Failure                  & S \\
  & Myocardial Infarction                     & S \\
  & Diabetes Mellitus                         & S \\
  & Chronic Obstructive Pulmonary Disease     & S \\
  & Stroke / Cerebrovascular Disease          & S \\

\bottomrule
\end{tabular}

\begin{tablenotes}
\scriptsize
\item[] \textit{Note:} T = Temporal (First 24~h of ICU stay); S = Static (single value per patient).
\item[a] GCS total computed as sum of eye, verbal, and motor scores.
\item[b] Vasoactive agents and mechanical ventilation encoded as per-hour binary indicators (1 = administered/active); comorbidities encoded from ICD-9/ICD-10 diagnoses.
\end{tablenotes}

\end{threeparttable}
\end{table}

\begin{figure*}[!t]
    \centering
    \includegraphics[width=1\linewidth]{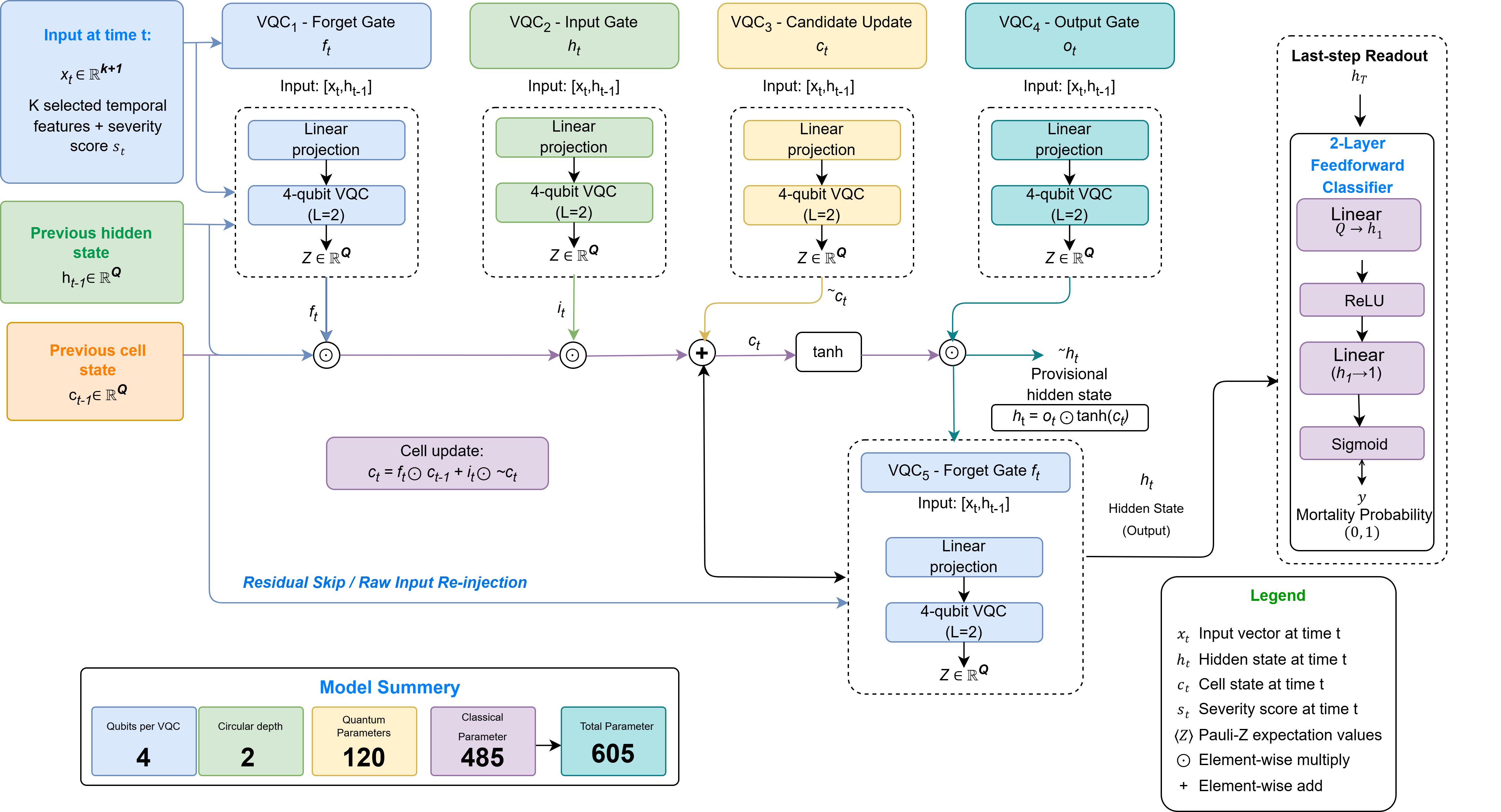}
    \caption{QuanTiMedAI methodology framework: the end-to-end pipeline covering cohort selection from MIMIC-IV, time-series preprocessing with temporal binning $T \in \mathcal{T}$, agentic Gemma 4 guided feature selection with iterative refinement, severity score computation, and the proposed 5-VQC QLSTM recurrent architecture with residual skip connection evaluated against classical LSTM baselines.}
    \label{fig:Methodological_Framework}
\end{figure*}

\subsection{Agentic Feature Selection via Large Language Model}

\subsubsection{Agent Design and Objective}
Feature selection was performed using an agentic framework
driven by a large language model (LLM), accessed locally
via Ollama. The agent operates in two phases: initial
feature selection and iterative performance-guided
refinement, with both phases restricted strictly to the
training set. At each configuration $(K, T)$, the
agent receives a mortality-stratified statistical summary
of the training set comprising feature-level
correlation with the outcome, mean $\pm$ SD stratified
by mortality status, temporal trend direction across the
first and second halves of the observation window, and
percentage missingness. The agent is instructed to return
exactly $K$ feature names along with a weight vector
$\mathbf{w} \in \mathbb{R}^{K}$ and a scalar bias $b$,
where
\begin{equation}
    K \in \mathcal{K},
    \label{eq:K}
\end{equation}
with $\mathcal{K}$ denoting the set of candidate feature
counts evaluated across experiments.

\subsubsection{Severity Score Engineering}
The selected features and LLM-derived weights are used to
construct an engineered severity score channel that is
appended to the input tensor at each temporal step. The
severity score at time $t$ is defined as:
\begin{equation}
    s(t) = \sigma\!\left(
        \sum_{i \in \mathcal{S}} w_i \, x_i(t) + b
    \right),
    \qquad
    \sigma(z) = \frac{1}{1 + e^{-z}},
    \label{eq:severity}
\end{equation}
where $\mathcal{S}$ denotes the set of selected features,
$w_i$ is the LLM-derived weight for feature $i$,
$x_i(t)$ is the normalized feature value at temporal
bin $t$, $b$ is a scalar bias term, and $\sigma(z)$
is the standard sigmoid activation function. This channel
provides the downstream model with a clinically informed
summary of patient deterioration at each time step,
embedding domain knowledge directly into the input
representation.

\subsubsection{Iterative Refinement Protocol}
Following initial selection, the feature set undergoes
up to $R$ rounds of iterative refinement. In each round,
the agent receives the current feature set, the derived
severity weights, and structured performance feedback
comprising fold-mean AUROC, AUPRC, cross-entropy loss,
and per-feature permutation importance scores.

To generate these feedback signals without data leakage,
a $V$-fold stratified cross-validation protocol is applied
exclusively to the training set.We kept the held-out test set completely separate during this optimization phase. Within each fold, we split the 
training data into training and validation sets using stratified sampling to keep the class balance even. We calculated feature standardization parameters using only the training split of each fold, then applied them to the validation split.On each fold, we trained an LSTM-based proxy network, using early stopping tied to the validation AUROC, and saved the best-performing checkpoint for fold evaluation. We then pooled the predictions across all folds to 
calculate our overall metrics, specifically the fold-mean $\pm$ SD and 95 %
confidence intervals for AUROC, AUPRC, and Brier score, alongside the average per-feature permutation importance.

To force the agent to keep exploring the feature space, we required it to swap out at least $\lceil\alpha \cdot K\rceil$ features from its current lineup at each refinement round (where $\alpha$ is the minimum swap fraction). Once this optimization grid locked in the feature selections and severity weights, all downstream experiments inherited them without needing to call the LLM again.

\subsection{Model Architecture}

All of our models share the same straightforward design: a single recurrent encoder layer, a last-step hidden state readout, and a two-layer feed-forward classifier with a sigmoid output. We intentionally kept this architecture minimal to isolate how much the recurrent core itself contributes to the performance.Each model takes an input tensor of shape $(N, T, K_{\text{in}})$, where $N$ is the batch size, $T$ is the number of temporal bins, and $K_{\text{in}}$ is the input feature dimensionality. For the LLM-guided variants, $K_{\text{in}} = K + 1$ to include the engineered severity score channel. For the random selection baseline, there is no extra channel, so $K_{\text{in}} = K$.

\begin{figure}[!t]
    \centering
    \includegraphics[width=1\columnwidth]{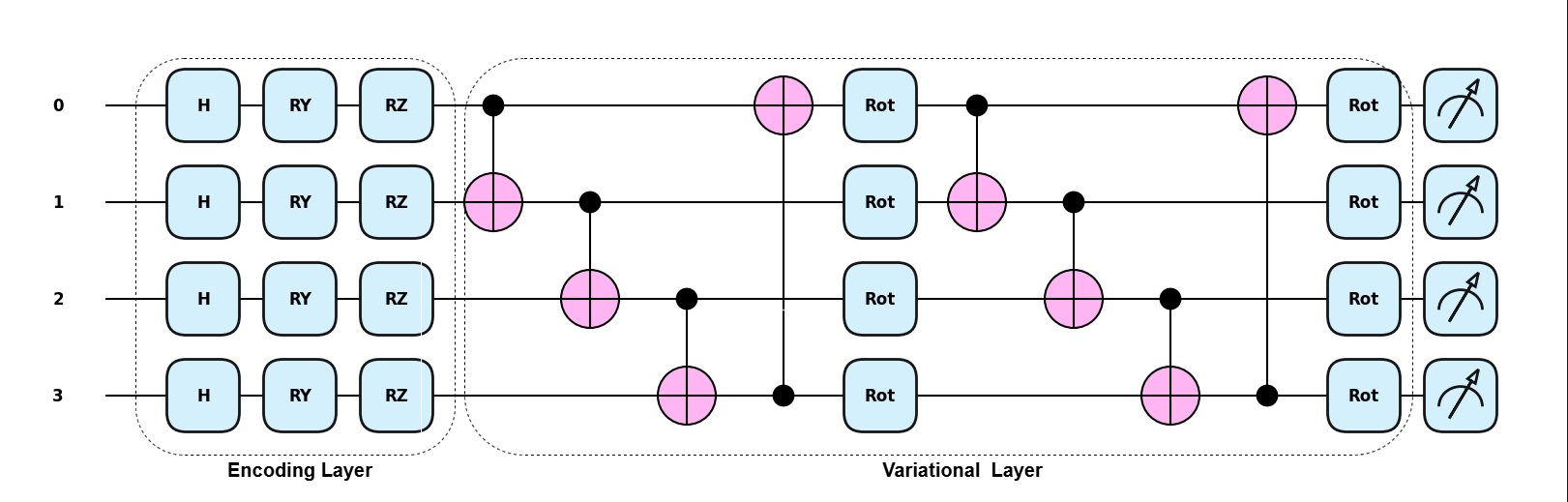}
    \caption{VQC Used in our proposed method}
    \label{fig:myCircuit}
\end{figure}

\subsubsection{Classical LSTM Baseline}
The classical baseline employs a two-layer stacked LSTM
architecture~\cite{hochreiter1997long}. Let $\mathbf{h}_t
\in \mathbb{R}^{d_h}$ and $\mathbf{c}_t \in
\mathbb{R}^{d_h}$ denote the hidden state and cell state
at temporal step $t$, respectively, where $d_h$ is the
hidden dimension. The standard LSTM transition at each
step is governed by:
\begin{align}
    \mathbf{f}_t &= \sigma(W_f \mathbf{x}_t +
        U_f \mathbf{h}_{t-1} + b_f), \notag \\
    \mathbf{i}_t &= \sigma(W_i \mathbf{x}_t +
        U_i \mathbf{h}_{t-1} + b_i), \notag \\
    \tilde{\mathbf{c}}_t &= \tanh(W_c \mathbf{x}_t +
        U_c \mathbf{h}_{t-1} + b_c), \notag \\
    \mathbf{c}_t &= \mathbf{f}_t \odot \mathbf{c}_{t-1}
        + \mathbf{i}_t \odot \tilde{\mathbf{c}}_t, \notag \\
    \mathbf{o}_t &= \sigma(W_o \mathbf{x}_t +
        U_o \mathbf{h}_{t-1} + b_o), \notag \\
    \mathbf{h}_t &= \mathbf{o}_t \odot
        \tanh(\mathbf{c}_t),
    \label{eq:lstm}
\end{align}
where $W_{\cdot}$, $U_{\cdot}$, and $b_{\cdot}$ are
learnable weight matrices and bias vectors, and $\odot$
denotes element-wise multiplication. The first layer
operates at hidden dimension $d_1$ and the second at
$d_2$, with the final hidden state $\mathbf{h}_T$ passed
to the classifier head.

\begin{figure*}[!t]
    \centering
    \includegraphics[width=1\textwidth]{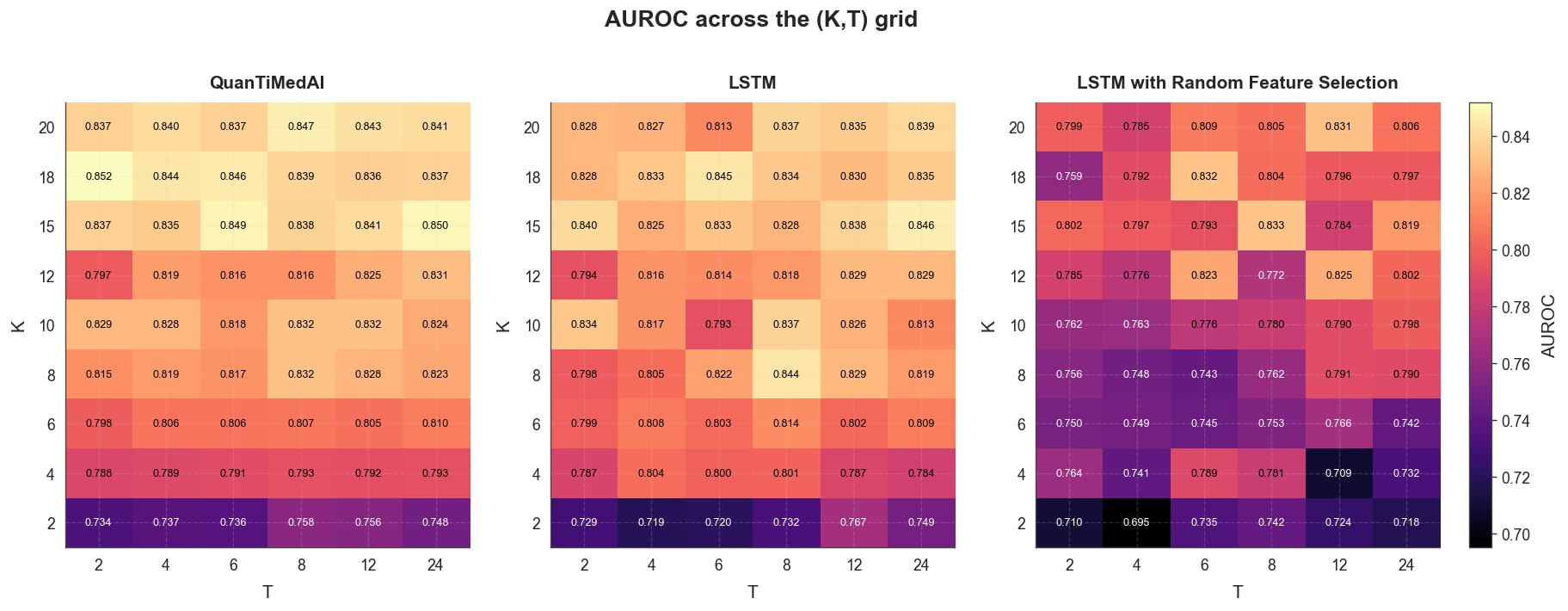}
    \includegraphics[width=1\textwidth]{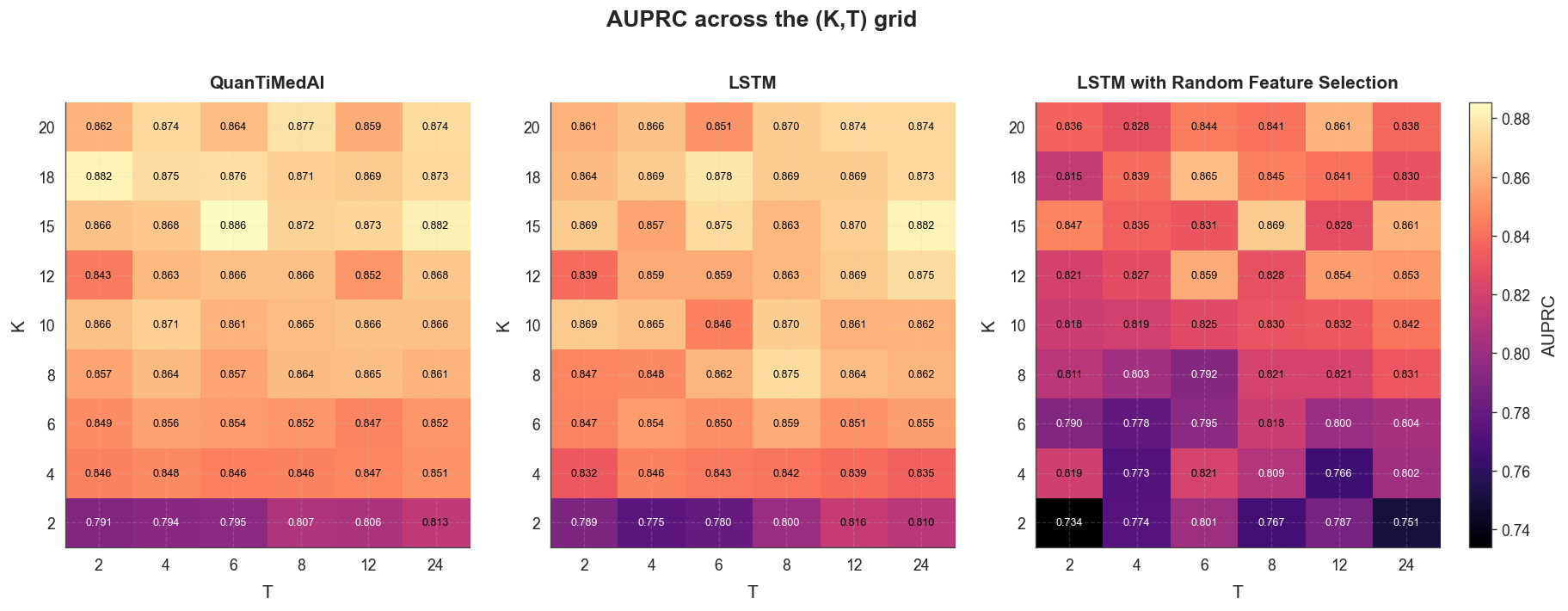}
    \caption{AUROC and AUPRC heatmaps for QuanTiMedAI(Ours), LSTM, and LSTM Random Feature Selection across the number of features $K$ and window $T$}
    \label{fig:Performance_Image}
\end{figure*}

\subsubsection{Proposed QLSTM Architecture}
The proposed QuanTiMedAI replaces each classical gate
with a hybrid quantum-classical block consisting of a
classical linear projection followed by a variational
quantum circuit (VQC)~\cite{chen2022qlstm,mitarai2018quantum}.
The cell contains $M$ VQCs distributed across the gate
structure: the forget gate employs $\text{VQC}_1$, the
input and update gates employ $\text{VQC}_2$ and
$\text{VQC}_3$ respectively, the output block employs
$\text{VQC}_4$, and the hidden-state refinementemploys $\text{VQC}_5$. The output-stage VQC
present in the original six-VQC
specification~\cite{chen2022qlstm} is omitted, reducing
the cell to five VQCs. A residual skip connection reinjects
the raw input $\mathbf{x}_t$ directly into the
hidden-refinement $\text{VQC}_5$, preserving clinical
signal that would otherwise be lost through the quantum
bottleneck.

Each VQC operates on $Q$ qubits. The circuit structure
consists of three stages. In the data encoding stage,
a Hadamard gate $H$ is applied to each qubit $q$ to
initialize a uniform superposition, followed by angle
encoding of the classical input. For each qubit $q$,
an RY rotation parameterized by $\arctan(x_q)$ and an
RZ rotation parameterized by $\arctan(x_q^2)$ are
applied:
\begin{equation}
    U_{\text{enc}}(x_q) =
        R_Z\!\left(\arctan(x_q^2)\right)
        R_Y\!\left(\arctan(x_q)\right) H,
    \label{eq:encoding}
\end{equation}
where $x_q$ denotes the projected input value for qubit
$q$. In the variational layer stage, $L$ variational
layers are applied. Within each layer, CNOT gates are
applied in a cyclic entanglement pattern connecting each
qubit to its neighbor, followed by a general rotation
gate $\text{Rot}(\phi_1, \phi_2, \phi_3)$ parameterized
by three independently trainable angles per qubit:
\begin{equation}
    U_{\text{var}}^{(\ell)} =
    \prod_{q=0}^{Q-1}
    \text{Rot}\!\left(
        \phi_q^{(\ell,1)},
        \phi_q^{(\ell,2)},
        \phi_q^{(\ell,3)}
    \right)
    \cdot
    \text{CNOT}_{\text{cyclic}},
    \label{eq:varlayer}
\end{equation}
where $\ell$ indexes the variational layer and
$\phi_q^{(\ell,j)}$ are the trainable rotation
parameters. In the measurement stage, the expectation
value of the Pauli-$Z$ operator is measured on each
qubit, yielding a classical output vector
$\mathbf{v} \in \mathbb{R}^{Q}$:
\begin{equation}
    v_q = \langle \psi | Z_q | \psi \rangle,
    \quad q = 0, \ldots, Q-1,
    \label{eq:measurement}
\end{equation}
where $|\psi\rangle$ denotes the final quantum state
after all circuit layers. All $M$ VQCs share this
structural design but are equipped with independently
trainable parameters, yielding a total of
$\Theta_Q = M \cdot Q \cdot L \cdot 3$ quantum
trainable parameters. The final hidden state
$\mathbf{h}_T$ is read out at the last temporal step
and passed to a classifier head. All VQCs are implemented
and executed using the \texttt{default.qubit} statevector
simulator in PennyLane~\cite{bergholm2018pennylane}, which
provides exact noiseless gradients and expectation values.
All reported results therefore reflect ideal simulation
conditions and do not account for gate errors, decoherence,
or readout noise present on physical quantum hardware.

\subsubsection{Implementation Details and Reproducibility}
To allow exact replication of the agentic feature selection pipeline and all downstream models, we report the full configuration used across experiments. Feature selection was handled by Gemma~4(E4B), served locally via Ollama under the identifier \texttt{gemma4:e4b}, with thinking mode enabled and reasoning traces stripped. The system prompt instructs the model to act as a senior intensivist and data scientist, reason solely from training-set summary statistics, and return a strict JSON object with keys \texttt{features}, severity weights, and severity bias, where at least one selected feature carries $|weight| \geq 0.10$. The initial-selection prompt supplies the mortality-stratified summary per feature, covering correlation with mortality, percentage missingness, dead/alive mean $\pm$ SD, and early-versus-late temporal trend, and instructs the agent to select exactly $K$ features with corresponding severity weights and bias. Each refinement prompt additionally supplies the previous feature set and severity parameters, the aggregate 10-fold inner cross-validation AUROC, AUPRC, and loss, the per-feature permutation-importance ranking, a table of the top-30 features by $|\text{correlation}|$, and from round~3 onward, the set of previously attempted combinations, which the agent is forbidden to repeat. Temperature was set to 0.30 for initial selection and 0.70 for refinement. Failed responses, defined as incorrect JSON schema, wrong $K$, duplicate features, or all-zero severity weights, triggered retries at temperature $\min(T_{\text{base}} + 0.15 \times \text{attempt},\, 1.20)$, with up to six retries on the initial call and eight on each refinement round. The refinement loop ran for exactly $R = 5$ rounds per $(K, T)$ configuration, with the agent required to swap at least a fraction $\alpha = 0.10$ of the $K$ selected features between consecutive rounds. For the headline configuration ($K = 18$), this enforces a minimum churn of $\lceil 0.10 \times 18\rceil = 2$ features per round; proposals violating this constraint, or numerically identical to the previous state, were rejected and resampled at higher temperature. The cross-validation feedback driving this loop used $V = 10$-fold stratified cross-validation restricted to the training partition. Within each fold, the training rows were further split 80/20 (stratified) into inner-training and inner-validation subsets used solely for early stopping, with z-score normalization parameters computed exclusively on the inner-training rows. Each fold trained an LSTM-based proxy for up to 100 epochs with early stopping (patience 20 epochs, monitored on validation AUROC), after which AUROC, AUPRC, cross-entropy loss, and permutation importance were computed on the fold's held-out rows. Fold-mean $\pm$ SD, a $t$-distribution-based 95\% confidence interval, and the pooled out-of-fold AUROC and AUPRC were then reported back to the agent as feedback.
\newtcolorbox{promptbox}[1][]{
  colback=gray!5,
  colframe=blue!50!black,
  fontupper=\small,
  breakable,
  title=#1
}

\begin{promptbox}[System Prompt]
You are a Senior Intensivist and Data Scientist specialising in post-cardiac-arrest ICU mortality. You are designing the input features for a time-series LSTM that predicts in-hospital death from the first 24 hours post-arrest.

\medskip
\textbf{Rules:}
\begin{itemize}
  \item You will ONLY ever be given summary statistics computed from training data — never any test-set information.
  \item Your recommendations are validated by cross-validation on that training data; downstream selection uses the CV results, not any held-out labels.
  \item Think carefully about which features have the strongest, most physiologically plausible relationship with mortality, penalising features with very high missingness or unstable scales.
  \item Reply STRICTLY in JSON with exactly these keys:
        \begin{itemize}
          \item \texttt{features} : list of feature-name strings
          \item \texttt{severity\_weights} : dict mapping feature-name → numeric weight (literals, not strings; in $[-6, 6]$)
          \item \texttt{severity\_bias} : numeric (not a string)
        \end{itemize}
  \item Emit NOTHING outside of that JSON object — no preamble, no epilogue, no markdown code fences if possible.
  \item At least one selected-feature weight must have $|\text{weight}| \geq 0.10$.
\end{itemize}

\end{promptbox}

\begin{promptbox}[Initial Selection Prompt]
Dataset: adult ICU patients post-cardiac arrest, first 24 hours.

\medskip
\textbf{POPULATION STATISTICS} (DEV set only — held-out test patients excluded):
\begin{center}
\texttt{\{dev\_labeled\_summary\}}
\end{center}

\textbf{TASK:}
\begin{enumerate}
  \item Select EXACTLY \texttt{\{K\}} features for the LSTM input. Prioritise large $|\text{correlation}|$ with mortality and clear dead/alive separation.
  \item Design a severity-score formula via \texttt{severity\_weights} and \texttt{severity\_bias}.
  \item Provide NUMERIC \texttt{severity\_weights} (not strings).
  \item At least one selected feature must have $|\text{weight}| \geq 0.10$.
\end{enumerate}

All available features: \texttt{\{all\_features\}}
\par
Return strictly JSON as specified in the system message.
\end{promptbox}

\begin{promptbox}[Refinement Prompt]
Previous features (\texttt{\{n\}}):
\begin{center}
\texttt{\{prev\_state['features']\}}
\end{center}
Previous severity weights: \texttt{\{prev\_state['severity\_weights']\}}
\par
Previous severity bias   : \texttt{\{prev\_state.get('severity\_bias', 0.0)\}}

\medskip
\textbf{INNER DEV-CV RESULTS} (dev-only, aggregate over \texttt{\{N\_CV\_FOLDS\}} folds):
\begin{itemize}
  \item AUROC = \texttt{\{inner\_cv\_metrics['auroc']:.3f\}}
  \item AUPRC = \texttt{\{inner\_cv\_metrics['auprc']:.3f\}}
  \item Loss  = \texttt{\{inner\_cv\_metrics['loss']:.3f\}}
\end{itemize}

\textbf{PERMUTATION IMPORTANCE} on inner-val folds (AUROC drop when shuffled):
\begin{center}
\texttt{\{imp\_lines\}}
\end{center}

\textbf{TOP-|CORRELATION| FEATURES} (dev-only, for context — not required):
\begin{center}
\texttt{\{corr\_lines\}}
\end{center}

\textbf{INNER-TRAIN STATISTICS} (dead vs alive, inner-train rows only):
\begin{center}
\texttt{\{inner\_fold\_train\_summary\}}
\end{center}

All available features: \texttt{\{all\_features\}}
\par
\texttt{\{forbidden\}}

\medskip
\textbf{TASK:} Propose a REFINED set of EXACTLY \texttt{\{K\}} features and updated \texttt{severity\_weights} and \texttt{severity\_bias}.
\begin{itemize}
  \item You MUST swap AT LEAST \texttt{\{n\_min\_swap\}} feature(s) vs the previous set.
  \item Drop features with the smallest permutation-importance drop.
  \item Consider adding high-$|\text{corr}|$ features that are NOT already selected.
  \item Provide NUMERIC \texttt{severity\_weights} (not strings). At least one selected-feature weight must have $|\text{weight}| \geq 0.10$.
  \item Keep weights and bias inside $[-6, 6]$.
\end{itemize}
Return strictly JSON as specified in the system message.
\end{promptbox}

 Once the five-round loop terminated, the incumbent with the highest mean inner CV AUROC, comprising the 18 selected features, their LLM-derived severity weights, and the severity bias listed in full in Table~\ref{tab:feature}, was frozen for the $K = 18$, $T = 2$ configuration and inherited unchanged by all downstream models trained at this cell, including the LSTM-Gemma baseline, every architectural ablation arm in Table~\ref{tab:ablation}, and QuanTiMedAI itself, which did not re-query the agent. QuanTiMedAI's recurrent core comprises a single QLSTM cell with five variational quantum circuits, covering the forget, input, candidate, output, and hidden-refinement gates, where the hidden-refinement circuit additionally receives the raw input $\mathbf{x}_t$ re-injected alongside the post-gate hidden state, each operating on $Q = 4$ qubits and simulated on PennyLane's noiseless \texttt{default.qubit} statevector backend with backpropagation differentiation, with a single linear projection preceding each circuit's data-encoding stage. All models were trained with the Adam optimizer at an initial learning rate of $1\times10^{-3}$ annealed via cosine decay to $1\times10^{-5}$ over 100 epochs, weight decay of $1\times10^{-4}$, gradient-norm clipping at 1.0, batch size 64, early stopping with patience 20 epochs on validation AUROC, and class-weighted binary cross-entropy loss to account for the cohort's mortality imbalance. Each model was retrained across three random seeds, with performance reported as seed-averaged test probabilities.

\section{Evaluation and Results}

\begin{table*}[!t]
\centering
\caption{Performance comparison of QuanTiMedAI with existing
         in-hospital cardiac arrest mortality prediction studies
         on MIMIC-IV.}
\label{tab:comparison_results}
\footnotesize
\setlength{\tabcolsep}{3pt}
\renewcommand{\arraystretch}{1.3}
\begin{tabular}{p{2cm} p{2.5cm} p{2.5cm} c c c}
\toprule
\textbf{Study} &
\textbf{Method} &
\textbf{Feature Selection} &
\textbf{Features} &
\textbf{AUROC} &
\textbf{Temp.} \\
\midrule
Sun et al.~\cite{sun2023prediction}
  & LASSO
  & LASSO
  & 11
  & 0.799
  & $\times$ \\

Li et al.~\cite{li2025prognostic}
  & RF
  & LASSO
  & 7
  & 0.830
  & $\times$ \\
Jia et al.~\cite{jia2026prediction}
  & XGBoost
  & Boruta
  & 37
  & 0.828
  & $\times$ \\
\textbf{QuanTiMedAI}
  & \textbf{Ours}
  & \textbf{Agentic LLM}
  & \textbf{18}
  & \textbf{0.852}
  & \checkmark \\
\bottomrule
\end{tabular}
\begin{tablenotes}
\footnotesize

\item[] Temp.\ = temporal modeling;
$\times$ = static; \checkmark = time-series.
\end{tablenotes}
\end{table*}

Table~\ref{tab:comparison_results} compares QuanTiMedAI against three recent in-hospital cardiac arrest mortality prediction studies on MIMIC-IV, summarizing each study's feature selection method, number of selected features, reported AUROC, and use of temporal modeling. Unlike these studies, which rely on static first-24-hour feature aggregation and conventional feature selection, QuanTiMedAI combines agentic LLM-guided feature selection, sequential time-series modeling, and a quantum-enhanced recurrent architecture. QuanTiMedAI's AUROC of 0.852 represents an improvement of approximately 2.9\% over the 0.828 reported by Jia et al.~\cite{jia2026prediction}, the most recent study from our literature review. 

For the controlled experiments that follow, we evaluate our proposed quantum LSTM (QLSTM) model against two classical LSTM baselines. One being a classical LSTM that uses the identical feature selection and severity weighting as our Model, provided by the Gemma language model. This isolates the architectural contribution of the quantum component. In another approach, classical LSTM with randomly selected feature subsets, representing a traditional black box approach without guided feature engineering.

All models are trained and assessed on the same MIMIC IV derived Cardiac arrest dataset for in-hospital mortality prediction. Inputs consist of time-series clinical variables.The number of features K is varied to study the impact of input dimensionality.

\subsection{Overall Performance Comparison}
Table~\ref{tab:performance_summary} aggregates the results across all 54 ($K$, $T$) configurations ($2 \le K \le 20$, $T \in \{2,4,6,8,12,24\}$~h). Our Model attains the highest mean AUROC (0.815), outperforming both the LSTM (0.810) and the random selection based LSTM (0.775). The same trend holds for AUPRC (0.856~vs~0.853~vs~0.819), demonstrating consistently superior discrimination and calibration. More importantly, the single best configuration of our model $K = 18$, $T = 2$~h reaches an AUROC of 0.852 and AUPRC of 0.882, markedly higher than the best LSTM setting (AUROC = 0.846, AUPRC = 0.882 at $K = 15$, $T = 24$~h) and the best random LSTM (AUROC = 0.833, AUPRC = 0.869).

\begin{table}[!t]
\caption{Performance Summary}
\label{tab:performance_summary}
\centering
\scriptsize
\setlength{\tabcolsep}{2.5pt}
\renewcommand{\arraystretch}{1.1}
\begin{tabular}{@{}l l c c c@{}}
\toprule
\textbf{Metric} & \textbf{Stat.} &
\makecell{\textbf{QuanTiMedAI}\\\textbf{}} &
\makecell{\textbf{LSTM}\\\textbf{}} &
\makecell{\textbf{LSTM}\\\textbf{Random Selection}} \\
\midrule
\multirow{3}{*}{\makecell[l]{\textbf{AUROC}\\$\uparrow$}}
  & Mean  & \textbf{0.815} & 0.810 & 0.775 \\
  & Best  & \makecell{\textbf{0.852}\\(K=18,T=2)} & \makecell{\textbf{0.846}\\(K=15,T=24)} & \makecell{\textbf{0.833}\\(K=15,T=8)} \\
  & Worst & \makecell{0.734\\(K=2,T=2)} & \makecell{0.719\\(K=2,T=4)} & \makecell{0.695\\(K=2,T=4)} \\
\midrule
\multirow{3}{*}{\makecell[l]{\textbf{AUPRC}\\$\uparrow$}}
  & Mean  & \textbf{0.856} & 0.853 & 0.819 \\
  & Best  & \makecell{\textbf{0.886}\\(K=15,T=6)} & \makecell{\textbf{0.882}\\(K=15,T=24)} & \makecell{\textbf{0.869}\\(K=15,T=8)} \\
  & Worst & \makecell{0.791\\(K=2,T=2)} & \makecell{0.775\\(K=2,T=4)} & \makecell{0.734\\(K=2,T=2)} \\
\midrule
\multicolumn{2}{@{}l}{\textbf{Best Configurations}}  & \textbf{K=18,T=2} & \textbf{K=15,T=24} & \textbf{K=15,T=8} \\

\bottomrule
\end{tabular}

\end{table}

\begin{table*}[htbp]
\centering
\caption{Parameter and performance comparison (K=18, T=2).}
\label{tab:combined}
\begin{tabular}{@{} l *{3}{S[table-format=2.0]} S[table-format=6.0] *{2}{S[table-format=6.0]} S[table-format=3.0] S[table-format=1.3] S[table-format=1.3] @{}}
\toprule
& \multicolumn{3}{c}{Settings} & \multicolumn{4}{c}{Parameters} & \multicolumn{2}{c}{Performance} \\
\cmidrule(lr){2-4} \cmidrule(lr){5-8} \cmidrule(lr){9-10}
\textbf{Experiment} & {K} & {T} & {Input Dim.} & {Total} & {Trainable} & {Quantum} & {Classical} & {AUROC} & {AUPRC} \\
\midrule
LSTM                    & 18 & 2 & 19 & 281729 & 281729 & 0   & 281729 & 0.828 & 0.863 \\
LSTM Random             & 18 & 2 & 18 & 280961 & 280961 & 0   & 280961 & 0.759 & 0.814 \\
\textbf{QuanTiMedAI(Ours)} & 18 & 2 & 19 & 605    & 605    & 120 & 485    & \textbf{0.852} & \textbf{0.882} \\
\bottomrule
\end{tabular}
\end{table*}

\subsection{ Parameter Efficiency}

A key advantage of our architecture is its drastically reduced parameter count. Table \ref{tab:combined} contrasts the model sizes for the $K = 18$, $T = 2$ setting. The classical LSTM requires 281,729 trainable parameters, while the random selection counterpart uses 280,961. Our quantum enhanced model operates with only 605 parameters total, 120 of which reside in the variational quantum circuit and 485 in the lightweight classical head. The most meaningful comparison is against a parameter-matched classical LSTM of 655 parameters, which achieves an AUROC of 0.835.QuanTiMedAI achieves 0.852 with a similar parameter budget, a gain of 0.017 AUROC (a relative improvement of approximately 2\%) that can be attributed to the quantum gating mechanism rather than model size. The ${\approx}466{\times}$ reduction in parameters relative to the full 281,729-parameter LSTM highlights the compactness of the proposed model while maintaining competitive predictive performance.

\begin{figure*}[!t] 
    \centering
    \includegraphics[width=1\textwidth]{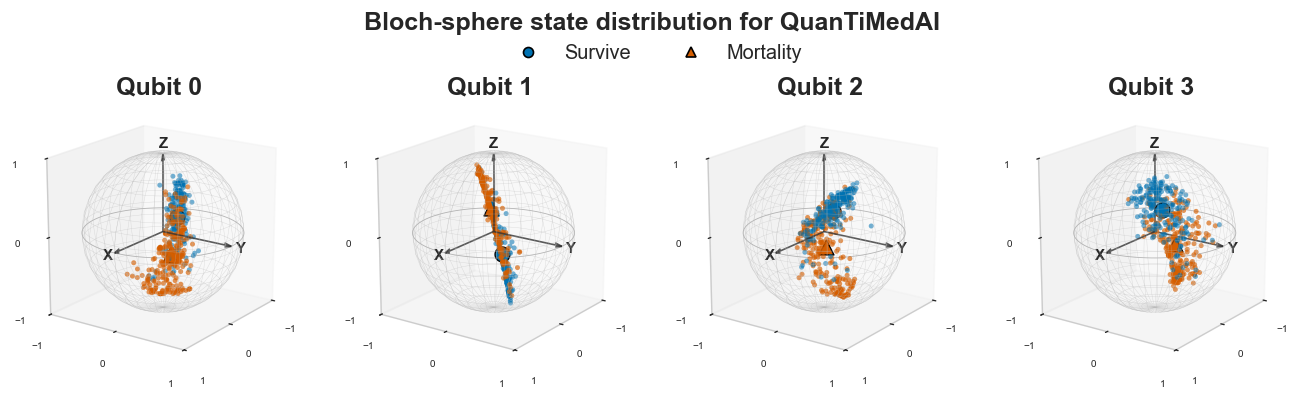}
    \caption{Bloch Sphere state distribution of VQC$_5$}
    \label{fig:Bloch_image}
\end{figure*}

\subsection{ Influence of Feature Count and Observation Window}
To understand the sensitivity to input design, we varied K  and T. As K grows from 2 to 18–20, all models improve substantially. Our Model consistently outperforms the baselines at every K, with the gap widening for intermediate K. At K = 10, our AUROC around 0.83 vs. 0.81 for LSTM. The top performance plateaus around K = 15–18, indicating that Gemma’s severity weighted feature ranking effectively captures the most discriminative physiological signals like lactate, base excess, anion gap, GCS, and SpO$_2$ being recurrently assigned the highest weights. Our model achieves its best result at T = 2 h, whereas the classical LSTM peaks at T = 24 h.

\subsection{Calibration and Robustness}
 While the pointwise improvements over the traditional LSTM are modest in a few configurations, our model consistently came out ahead across all 54 $(K, T)$ settings, including the mean, best, and worst cases. This uniform performance gap suggests a genuine, reproducible advantage rather than a statistical fluke.

\subsection{Learned Feature Severity Weights}
The Gemma-derived severity weights add a layer of interpretability to our model. In every setup, the largest positive coefficients belonged to lactate, anion gap, base excess, bicarbonate, GCS, and $\text{SpO}_2$, matching established clinical knowledge regarding mortality risk. Meanwhile, the bias term automatically adapts the model's decision threshold. This also implicitly compensates for unknown factors. Our model preserves these weights exactly as provided, demonstrating that the quantum component enhances the model without distorting the clinically meaningful feature attribution.

\subsection{Quantum State Trajectory and Bloch Sphere Analysis}

Figure~\ref{fig:Bloch_image} illustrates the final distributions across all four qubits. Each qubit develops a specialized geometric signature, proving the network distributes feature processing efficiently. This multi-axial separation confirms that QuanTiMedAI successfully exploits Hilbert space expressivity for robust linear separability.

\begin{figure*}[!t] 
    \centering
    \includegraphics[width=1\textwidth]{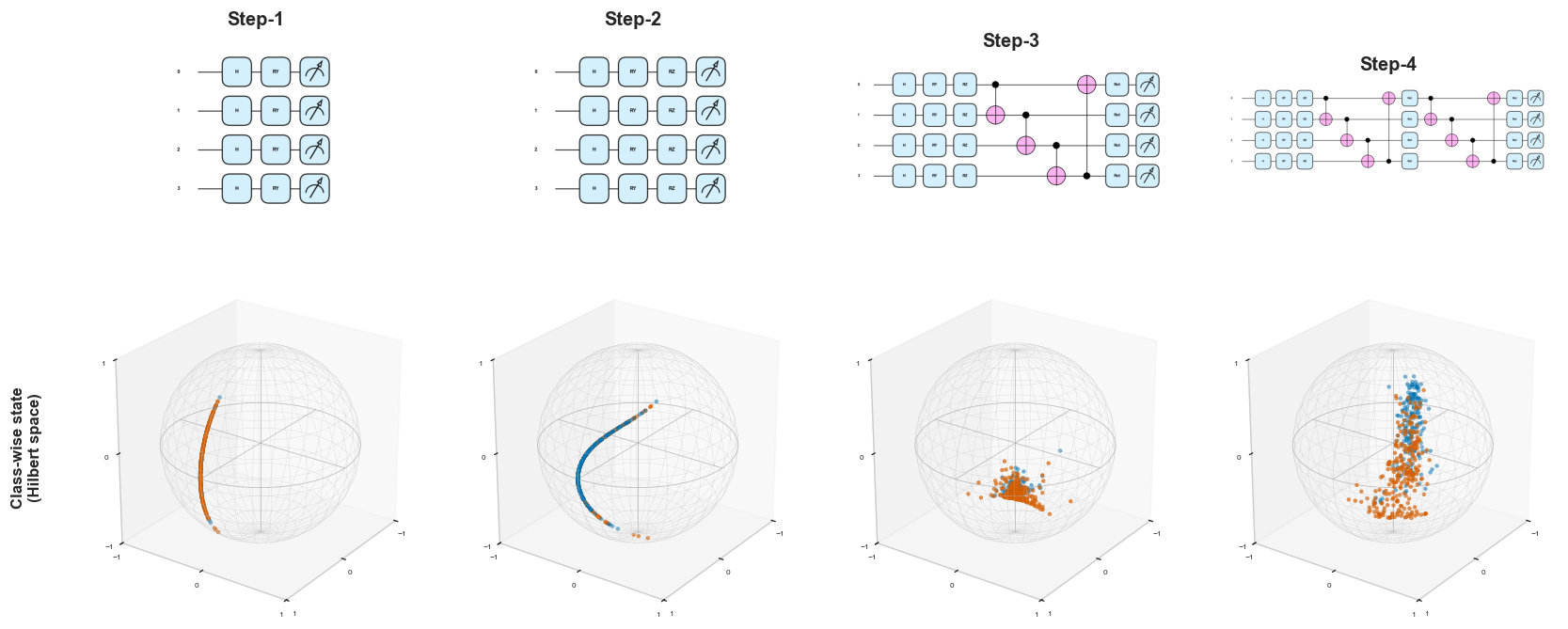}
    \caption{Quantum State Distribution of Qubit 0 as circuit grows at each step where blue means survived and yellow means mortality}
    \label{fig:Cir_image}
\end{figure*}

To evaluate how the Variational Quantum Circuit (VQC) transforms features into class-discriminative states, we analyze Bloch sphere trajectories. Figure~\ref{fig:Cir_image} tracks this evolution for Qubit 0. In Step-1 and Step-2, embedding gates ($H$, $R_Y$, $R_Z$) restrict pure states to the sphere's surface, leaving the \textit{Survive} and \textit{Mortality} classes completely interleaved.

A phase transition occurs in Step-3; entangling CNOT gates contract the state distributions toward the center, signaling mixed states induced by multi-qubit entanglement. By Step-4, variational layers leverage this space to repel states toward the boundaries, fracturing the data into two distinct, spatially segregated clusters.

\section{Ablation Study}

The original QLSTM architecture introduced by Chen et al.~\cite{chen2022qlstm}
employs six VQCs per recurrent cell, corresponding to the forget, input, cell
candidate, output, hidden-state refinement, and output-stage gates, with four
qubits and circuit depth two. This specification has since been adopted in
subsequent QLSTM applications. Khan et al.~\cite{khan2024qlstm} applied this
architecture to solar power forecasting, comparing QLSTM against classical LSTM
on real-world photovoltaic datasets and demonstrating accelerated training
convergence and reduced test loss within the initial epochs for QLSTM,
suggesting its capacity to capture complex spatiotemporal patterns more
efficiently than classical counterparts. Kea et al.~\cite{kea2024hybrid}
similarly demonstrated that a hybrid QLSTM outperformed classical LSTM on stock
price prediction across multiple evaluation metrics. While these studies
validate the original 6-VQC specification on structured time-series data, none
have investigated whether modifications to the VQC count, re-injection
mechanism, or temporal read-out strategy affect performance on high-dimensional
clinical EHR sequences. This work directly investigates that question.

\begin{table*}[!t]
\caption{QuanTiMedAI Architectural Ablation at $K{=}18$, $T{=}2$}
\label{tab:ablation}
\centering
\small
\setlength{\tabcolsep}{3pt}
\resizebox{\textwidth}{!}{%
\begin{tabular}{lccccccccccc}
\toprule
\textbf{Configuration} & \textbf{VQCs} & \textbf{Skip} & \textbf{Read-out} & \textbf{Total Params} & \textbf{Q-Params} & \textbf{Acc.} & \textbf{Prec.} & \textbf{Rec.} & \textbf{F1} & \textbf{AUROC} & \textbf{AUPRC} \\
\midrule
\textbf{QuanTiMedAI (Ours)} & \textbf{5} & \checkmark & last & 605 & 120 & 0.758 & 0.808 & 0.746 & 0.776 & \textbf{0.852} & \textbf{0.882} \\
6-VQC with Skip & 6 & \checkmark & last & 725 & 144 & \textbf{0.768} & 0.810 & 0.769 & 0.789 & 0.848 & 0.873 \\
5-VQC without Skip & 5 & --- & last & 529 & 120 & 0.766 & 0.814 & 0.758 & 0.785 & 0.839 & 0.868 \\
Mean-pool read-out & 5 & \checkmark & mean & 605 & 120 & 0.766 & \textbf{0.817} & 0.754 & 0.784 & 0.838 & 0.865 \\
Chen et al.~\cite{chen2022qlstm} & 6 & --- & last & 573 & 144 & 0.766 & 0.797 & \textbf{0.785} & \textbf{0.791} & 0.838 & 0.867 \\
Param.\ matched classical LSTM & --- & --- & last & 655 & 0 & 0.753 & 0.797 & 0.754 & 0.775 & 0.835 & 0.868 \\
Classical LSTM & --- & --- & last & 281{,}729 & 0 & 0.738 & 0.803 & 0.708 & 0.753 & 0.828 & 0.864 \\
\bottomrule
\end{tabular}%
}
\end{table*}

All ablation experiments were locked to $K{=}18$, $T{=}2$, inheriting the same
feature set, severity weights, and severity bias selected by the Gemma-4 agentic
pipeline at this configuration. The held-out test set ($N{=}462$,
mortality\,=\,56.3\%) was never used during feature or model selection. Each arm was trained across three seeds. Qubit count (four) and circuit depth
(two) were held constant across all quantum arms to isolate specific design
choices.

Six architectural variants were evaluated in a structured design. The first four arms form a $2{\times}2$ factorial over two design axes; VQC
topology (5-VQC vs.\ 6-VQC) and $x_t$ re-injection (on vs.\ off), so each
main effect is estimable both in isolation and in combination. The proposed
QuanTiMedAI occupies the 5-VQC with re-injection cell of this factorial; the
remaining three cells isolate what happens when either or both design choices
are removed. Two further arms probe the temporal read-out strategy and the
value of the quantum core itself.

The proposed architecture, \textbf{QuanTiMedAI}, is our modified QLSTM design
that removes the output-stage VQC from the original Chen et al.\ specification,
reducing the cell to five VQCs, and adds a residual skip connection that
reinjects the raw current input $x_t$ into the hidden-refinement VQC. The
temporal summary fed to the classifier head is the last hidden state $h_T$. The
remaining five arms each change exactly one design choice: \textbf{5-VQC
without re-injection} isolates the contribution of the skip connection; the
\textbf{Chen et al.~\cite{chen2022qlstm} baseline} retains the original
output-stage VQC with no skip connection and serves as the direct literature
baseline; \textbf{6-VQC with skip connection} completes the $2{\times}2$
factorial; \textbf{mean-pool read-out} replaces the last-step $h_T$ with the
mean of $h_t$ across all $T$ time steps; and the \textbf{classical LSTM} arm
replaces the quantum cell with a standard two-layer LSTM of hidden sizes 192
and 96 with last-step readout, serving as the classical upper-bound baseline just as it was done previously.

\begin{table*}[htbp]
\centering
\caption{Statistical comparison of baseline models against QuanTiMedAi ( $\text{AUROC} = 0.852$).}
\label{tab:model_comparison}
\small
\begin{tabular}{lcccccc}
\toprule
\textbf{Model} & \textbf{AUROC} & \textbf{$\Delta$AUROC } & \textbf{$z$} & \textbf{$p$ (raw)} & \textbf{$p$ (Bonf.)} & \textbf{Sig. @ 0.05} \\
\midrule
5-VQC without skip connection                                   &   0.839& 0.0127 & 2.089 & 0.0367  & 0.7711 & no  \\
Chen 2020: 6-VQC, no skip connection                        & 0.838 & 0.0142 & 2.619 & 0.0088  & 0.1851 & no  \\
6-VQC with skip connection                              & 0.848 & 0.0040 & 0.954 & 0.3403  & 1.0000 & no  \\
Mean-pool readout (vs last step)                        & 0.838 & 0.0135 & 2.944 & 0.0032  & 0.0680 & no  \\
No quantum core (classical LSTM)                        & 0.828 & 0.0236 & 2.492 & 0.0127  & 0.2667 & no  \\
Parameter matched classical LSTM & 0.835 & 0.0170 & 3.337 & $<0.001$ & 0.0178 & \textbf{yes} \\
\bottomrule
\end{tabular}
\end{table*}

The full capacity classical LSTM attains an AUROC of just 0.828, while a drastically smaller parameter-matched classical LSTM reaches 0.835. This inversion where a model with more parameters performs worse is a classic signature of overfitting on the limited training set. While the larger LSTM tends to memorize spurious patterns which hurts its ability to generalize the compact classical model avoids this issue entirely. However, the classical model's simpler gates ultimately bottleneck its representational power. QuanTiMedAI addresses both challenges at once. Within a similar parameter budget, it achieves an AUROC of 0.852, noticeably outperforming the parameter-matched classical LSTM (Bonferroni-corrected $p = 0.018$).This edge suggests that the quantum gating mechanism offers advantages beyond simple parameter reduction, though this benefit needs further validation on noisy simulators and physical hardware before making definitive claims. These results align with the idea that quantum entanglement can support more compact temporal encoding, even if the exact mechanism and its resilience to realistic noise remain open questions. QuanTiMedAI also surpasses the large LSTM by a margin of 0.024 AUROC, and while the raw 
p-value of 0.013 does not survive the most conservative correction, this is an expected power limitation given the test-set size, not evidence of equivalence. In contrast to classical models that either overfit (large) or underfit (small), QuanTiMedAi delivers the best of both compact size and inherent robustness to overfitting making it the clinically safer and more reliable choice

Table~\ref{tab:ablation} presents the Test results for all six arms,
ranked by AUROC.

\begin{table}[htbp]
\centering
\caption{Feature weights ($K=18$, $T=2$).}
\label{tab:feature}
\begin{tabular}{clS[table-format={+1.4}]}
\toprule
\# & Feature & \multicolumn{1}{c}{Severity weight} \\
\midrule
1 & Lactate & +0.6500 \\
2 & Base excess & +0.5500 \\
3 & Anion gap & +0.3500 \\
4 & Bicarbonate & +0.2000 \\
5 & Glasgow Coma Scale (GCS) & +0.3000 \\
6 & Oxygen saturation (SpO\(_2\)) & +0.2000 \\
7 & pH & +0.1000 \\
8 & Albumin & +0.2500 \\
9 & Systolic blood pressure (SBP) & -0.1500 \\
10 & Creatinine & +0.1000 \\
11 & Blood urea nitrogen (BUN) & +0.1000 \\
12 & Sodium & +0.0800 \\
13 & Heart rate & +0.0800 \\
14 & Respiratory rate & +0.0500 \\
15 & Glucose & +0.0500 \\
16 & Red cell distribution width (RDW) & +0.0500 \\
17 & Hemoglobin & +0.0500 \\
18 & Age & +0.0500 \\
\midrule
\multicolumn{2}{l}{Severity bias} & +0.5000 \\
\bottomrule
\end{tabular}
\end{table}

\begin{table}[htbp]
\centering
\caption{Performance metrics of QuanTiMedAI as Qubit count is increased}
\label{tab:qnn_perf}
\begin{tabular}{@{}l S[table-format=4.0] S[table-format=3.0] S[table-format=1.3] S[table-format=1.3]@{}}
\toprule
\textbf{Qubits} & \textbf{Total params} & \textbf{Quantum params} & \textbf{AUROC} & \textbf{AUPRC} \\
\midrule
2 qubits  & 283  & 60  & 0.843 & 0.872 \\
3 qubits  & 439  & 90  & 0.845 & 0.878 \\
4 qubits  & 605  & 120 & 0.852 & 0.882 \\
6 qubits  & 967  & 180 & 0.850 & 0.883 \\
8 qubits  & 1369 & 240 & 0.852 & 0.877 \\
\bottomrule
\end{tabular}
\end{table}

QuanTiMedAI, the proposed architecture, achieves the highest AUROC of 0.852, outperforming
all other arms. Comparing the two 5-VQC arms shows that $x_t$ re-injection
contributes a 0.013 AUROC improvement (approximately 1.5\% relative). This supports the architectural motivation:
reinjecting $x_t$ allows the model to recover signal lost through the 4-qubit
bottleneck, which is particularly consequential in ICU EHR sequences where
physiological measurements at each time step carry direct prognostic value.

Comparing QuanTiMedAI against the Chen et al.~\cite{chen2022qlstm} baseline 
the original specification used in prior QLSTM literature shows a gap of
0.014 AUROC in favour of the proposed architecture. The 6-VQC with re-injection
arm (AUROC\,0.848) sits between the two, confirming that re-injection is
consistently beneficial regardless of VQC count, while QuanTiMedAI remains the
most parameter-efficient high-performing configuration. Replacing the last-step
readout with mean pooling reduces AUROC from 0.852 to 0.838, confirming that
the final hidden state $h_T$ carries more discriminative information than a
uniform average of hidden states for ICU mortality prediction.

Taken together, the two classical baselines reveal a clear trend. The full-capacity LSTM (281,729 parameters) scores an AUROC of 0.828, while the parameter-matched version (655 parameters) actually hits 0.835. This minimal difference indicates that the larger model's massive capacity doesn't improve discrimination on this cohort. Instead, the dataset likely cannot support over 280k weights without overfitting; simply scaling the network down to a few hundred parameters yields comparable results.In contrast, QuanTiMedAI reaches an AUROC of 0.852 with just 605 parameters, outperforming both classical setups across every configuration. The jump over the parameter-matched LSTM (0.852 vs. 0.835) is especially telling. It confirms that the quantum gating mechanism itself, rather than just a lower parameter count drives the performance. While the resulting $\approx$466-fold compression over the full LSTM is a massive structural advantage, these findings demonstrate highly parameter-efficient performance under ideal simulation. We attribute this edge to the entangled variational circuit, though this requires validation under realistic noise conditions.

\section{Discussion}


QuanTiMedAI outperforms a comparably sized classical LSTM because of how 
variational quantum circuits handle clinical data. A classical model needs 
many parameters to simultaneously capture the nonlinear relationships 
between variables like lactate, GCS, pH, and base excess. VQCs do this 
more efficiently: by encoding these measurements into quantum state space 
and entangling the qubits, the circuit naturally picks up complex 
interactions across multiple physiological signals at once, without 
needing a large parameter 
budget~\cite{havlicek2019quantum,abbas2021expressibility}. This is why 
QuanTiMedAI achieves an AUROC of 0.852 with only 605 parameters while a 
comparably sized classical LSTM of 655 parameters reaches only 0.835, 
the quantum gating mechanism provides a richer representational capacity, 
not just a smaller footprint. The $x_t$ skip connection adds to this by 
keeping raw clinical measurements directly accessible at each time step, 
recovering signal that would otherwise be lost through the qubit 
bottleneck~\cite{laskar2025shallow}. Validation under realistic hardware 
noise remains an important next step.

\section{Conclusion}

In this study, we proposed QuanTiMedAI, a quantum-agentic
time-series framework that combines agentic LLM-guided feature
selection with a compact Quantum Long Short-Term Memory
architecture for in-hospital cardiac arrest mortality prediction
on MIMIC-IV. Agentic Gemma-4 feature selection consistently
outperformed random selection, and a structured ablation study
showed that the proposed 5-VQC architecture with input re-injection performed best among all variants, including the original Chen et al.~\cite{chen2022qlstm} specification, with re-injection contributing the largest single gain in both AUROC and calibration. QuanTiMedAI achieved an AUROC of 0.852 using only 605 parameters, compared to 281{,}729 for a full-capacity LSTM,
exceeding a parameter-matched classical LSTM (0.835) and improving
on a recent MIMIC-IV cardiac arrest baseline by approximately 2.9\%~\cite{jia2026prediction}.
These results support QuanTiMedAI as a lightweight candidate for future resource-aware clinical applications, pending external
validation.

\section*{REFERENCES}

\bibliographystyle{plainnat}

\bibliography{references}

\end{document}